\documentclass{article}
\usepackage{ijcai17}
\usepackage{times}
\usepackage{helvet}
\usepackage{courier}
\usepackage{color}
\usepackage{subfigure}
\usepackage{subfig}
\usepackage{graphicx}
\usepackage{bm}
\usepackage{subfig}
\usepackage{amsmath,amssymb}
\usepackage{graphics}
\usepackage{algorithm}
\usepackage{algorithmic}
\usepackage{url}
\usepackage{balance}
\usepackage{amsthm}
\usepackage{booktabs}

\title{Sparse Weighted Canonical Correlation Analysis
\thanks{Corresponding author.}
}

\author{Wenwen Min$^{1,2,3}$, Juan Liu$^{1,*}$ and Shihua Zhang$^{2,3,*}$\\
                  $^1$ State Key Laboratory of Software Engineering, School of Computer, Wuhan University\\
                  $^2$ National Center for Mathematics and Interdisciplinary Sciences, \\Academy of Mathematics and Systems Science, Chinese Academy of Sciences\\
                  $^3$ School of Mathematics Sciences, University of Chinese Academy of Sciences\\
                  \texttt{mww@whu.edu.cn, liujuan@whu.edu.cn and zsh@amss.ac.cn}}

\begin{document}
\maketitle
\begin{abstract}
Given two data matrices $\bm{X}$ and $\bm{Y}$, Sparse canonical correlation analysis (SCCA) is to seek two sparse canonical vectors $\bm{u}$ and $\bm{v}$ to maximize the correlation between $\bm{X}\bm{u}$ and $\bm{Y}\bm{v}$. However, classical and sparse CCA models consider the contribution of all the samples of data matrices and thus cannot identify an underlying specific subset of samples. To this end, we propose a novel Sparse weighted canonical correlation analysis (SWCCA), where weights are used for regularizing different samples. We solve the $L_0$-regularized SWCCA ($L_0$-SWCCA) using an alternating iterative algorithm. We apply $L_0$-SWCCA to synthetic data and real-world data to demonstrate its effectiveness and superiority compared to related methods. Lastly, we consider also SWCCA with different penalties like LASSO (Least absolute shrinkage and selection operator) and Group LASSO, and extend it for integrating more than three data matrices.
\end{abstract}
\section{Introduction}
Canonical correlation analysis (CCA) is a powerful tool to integrate two data matrices \cite{klami2013bayesian,sun2008least,yang2017canonical,cai2016kernel,wang2017novel}, which has been comprehensively used in many diverse fields. Given two matrices $\bm{X}\in \mathbb{R}^{n\times p}$ and $\bm{Y}\in \mathbb{R}^{n\times q}$ from the same samples, CCA is used to find two sparse canonical vectors $\bm{u}$ and $\bm{v}$ to maximize the correlation between $\bm{X}\bm{u}$ and $\bm{Y}\bm{v}$. However, in many real-world problems like those in bioinformatics \cite{witten2009penalized,mizutani2012relating,le2009sparse,fang2016joint,yoshida2017sparse}, the number of variables in each data matrix is usually much larger than the sample size. The classical CCA leads to non-sparse canonical vectors which are difficult to interpret in biology. To conquer this issue, a large number of sparse CCA models \cite{witten2009penalized,mizutani2012relating,le2009sparse,fang2016joint,yoshida2017sparse,parkhomenko2009sparse,witten2009extensions,asteris2016simple,chu2013sparse,hardoon2011sparse,gao2015minimax} have been proposed by using regularized penalties (\emph{e.g.}, LASSO and $L_0$-norm) to obtain sparse canonical vectors for variable selection. Parkhomenko \emph{et al.} \cite{parkhomenko2009sparse} first proposed a Sparse CCA (SCCA) model using LASSO ($L_1$-norm) penalty to genomic data integration. $\mbox{L}\hat{e}$ Cao \emph{et al.} \cite{le2009sparse} further proposed a regularized CCA with Elastic-Net penalty for a similar task. Witten \emph{et al.} \cite{witten2009penalized} proposed the Penalized matrix decomposition (PMD) algorithm to solve the Sparse CCA with two penalties: LASSO and Fused LASSO to integrate DNA copy number and gene expression from the same samples/individuals. Furthermore, a large number of generalized LASSO regularized CCA models have been proposed to consider prior structural information of variables \cite{lin2013group,virtanen2011bayesian,chen2012structure,chen2012structured,du2016structured}. For example, Lin \emph{et al.} \cite{lin2013group} proposed a Group LASSO regularized CCA to explore the relationship between two types of genomic data sets. If we consider a pathway as a gene group, then these gene pathways form an overlapping group structure \cite{jacob2009group}. Chen \emph{et al.} \cite{chen2012structured} developed an overlapping group LASSO regularized CCA model to employ such group structure.

These existing sparse CCA models can find two sparse canonical vectors with a small subset of variables across all samples (Fig.1(a)). However, many real data such as the cancer genomic data show distinct heterogeneity \cite{dai2015breast,chen2016integrative}. Thus, the current CCA models fail to consider such heterogeneity and cannot directly identify a set of sample-specific correlated variables. To this end, we propose a novel Sparse weighted CCA (SWCCA) model, where weights are used for regularizing different samples with a typical penalty (\emph{e.g.}, LASSO and $L_0$-norm) (Fig.1(b)). In this way, SWCCA can not only select two variable sets, but also select a sample set (Fig.1 (b)). We further adopt an efficient alternating iterative algorithm to solve $L_0$ (or $L_1$) regularized SWCCA model. We apply $L_0$-SWCCA and related ones onto two simulated datasets and two real biological data to demonstrate its efficiency in capturing correlated variables across a subset of samples.
\begin{figure}[ht]
  \centering
  \includegraphics[width=1\linewidth]{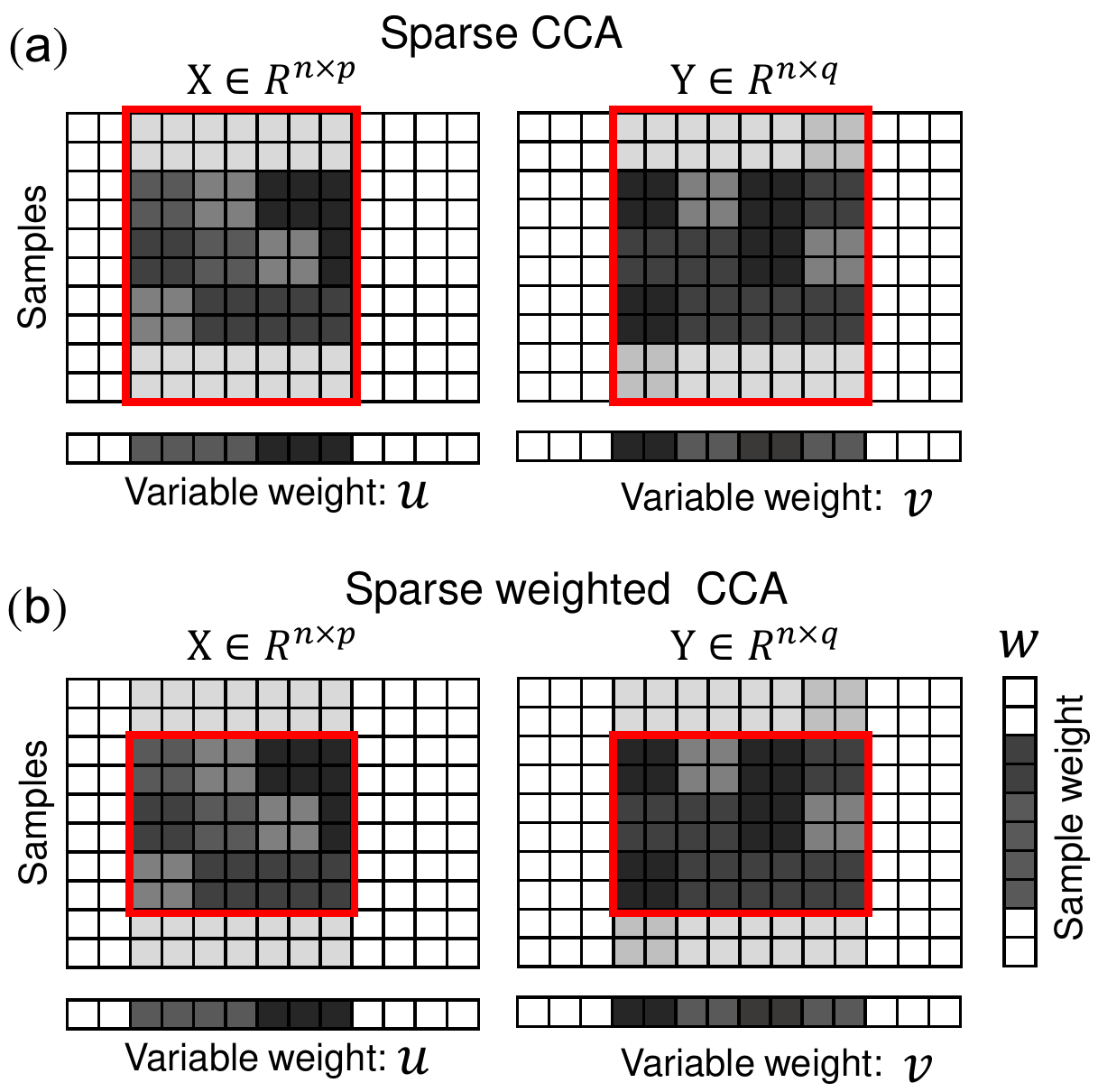}
  \caption{Illustration of the difference between SWCCA and SCCA. (a) SCCA is used to extract two sparse canonical vectors ($\bm{u}$ and $\bm{v}$) to measure the association of two matrices; (b) SWCCA is used to consider two subset of sample-related sparse canonical vectors. The weights ($\bm{w}$) are used for regularizing different samples  in SWCCA. SWCCA can not only obtain two sparse canonical vectors, but also identify a set of samples based on those non-zero elements of $\bm{w}$.}
\end{figure}
\section{$\bm{L_0}$-regularized SWCCA}
Here, we assume that there are two data matrices $\bm{X}\in \mathbb{R}^{n\times p}$ ($n$ samples and $p$ variables) and $\bm{Y} \in \mathbb{R}^{n\times q}$ ($n$ samples and $q$ variables) across a same set of samples. The classical CCA seeks two components ($\bm{u}$ and $\bm{v}$) to maximize the correlation between linear combinations of variables from the two data matrices as Eq.(1).
\begin{equation}
 \rho = \frac{\bm{u}^{\rm{T}} \bm{\Sigma}_{xy}\bm{v}}{\sqrt{(\bm{u}^{\rm{T}}\bm{\Sigma}_{x}\bm{u})(\bm{v}^{\rm{T}}\bm{\Sigma}_{y}\bm{v})}}
\end{equation}
If $\bm{X}$ and $\bm{Y}$ are centered, we obtain the empirical covariance matrices $\bm{\Sigma}_{xy} = \frac{1}{n}\bm{X}^{\rm{T}}\bm{Y}$, $\bm{\Sigma}_{x} = \frac{1}{n}\bm{X}^{\rm{T}}\bm{X}$ and $\bm{\Sigma}_{y} = \frac{1}{n}\bm{Y}^{\rm{T}}\bm{X}$. Thus we have the following equivalent criterion as Eq.(2).
\begin{equation}
 \rho = \frac{\bm{u}^{\rm{T}}\bm{X}^{\rm{T}}\bm{Y}\bm{v}} {\sqrt{(\bm{u}^{\rm{T}}\bm{X}^{\rm{T}}\bm{X}\bm{u})(\bm{v}^{\rm{T}}\bm{Y}^{\rm{T}}\bm{Y}\bm{v})}}
\end{equation}
Obviously, $\rho$ of (2) is invariant to the scaling of $\bm{u}$ and $\bm{v}$. Thus, maximizing criterion (2) is equivalent to solve the following constrained optimization problem as Eq.(3).
\begin{equation}
  \begin{array}{rl}
    \max_{\bm{u},\bm{v}} & \bm{u}^{\rm{T}}\bm{X}^{\rm{T}}\bm{Y}\bm{v} \\
    \mbox{s.t.}& \bm{u}^{\rm{T}}\bm{X}^{\rm{T}}\bm{X}\bm{u} =  1, \bm{v}^{\rm{T}}\bm{Y}^{\rm{T}}\bm{Y}\bm{v} = 1
  \end{array}
\end{equation}
Previous studies \cite{witten2009penalized,witten2009extensions} have shown that considering the covariance matrix ($\frac{1}{n}\bm{X}^{\rm{T}}\bm{X}$, $\frac{1}{n}\bm{Y}^{\rm{T}}\bm{Y}$) as diagonal one can obtain better results. For this reason, Asteris \emph{et al.} \cite{asteris2016simple} assume that $\bm{X}^{\rm{T}}\bm{X} = \bm{I}$ and $\bm{Y}^{\rm{T}}\bm{Y} = \bm{I}$, and the $L_0$-regularized Sparse CCA ($L_0$-SCCA) (also called ``diagonal penalized CCA'') model can be presented as Eq.(4).
\begin{equation}
  \begin{array}{rl}
    \max_{\bm{u},\bm{v}} & \bm{u}^{\rm{T}}\bm{X}^{\rm{T}}\bm{Y}\bm{v}\\
    \mbox{s.t.}& \|\bm{u}\|_0\leq k_u, \|\bm{v}\|_0\leq k_v\\
               & \bm{u}^{\rm{T}}\bm{u} = \bm{v}^{\rm{T}}\bm{v} = 1
  \end{array}
\end{equation}
where $\|\bm{u}\|_0$ is the $L_0$-norm penalty function, which returns to the number of non-zero entries of $\bm{u}$.
Asteris \emph{et al.}\cite{asteris2016simple}  applied a projection strategy to solve $L_0$-SCCA. Let $\bm{A}=\bm{X}^{\rm{T}}\bm{Y}$, then the model of Eq.(4) is equivalent to rank-one $L_0$-SVD model \cite{min2015novel}.

Let $\bm{a} = \bm{X}\bm{u}$ and $\bm{b} = \bm{Y}\bm{v}$, then the objective function $\bm{u}^{\rm{T}}\bm{X}^{\rm{T}}\bm{Y}\bm{v} = \sum_{i=1}^n a_ib_i$. To consider the different contribution for samples, we modify the objective function of Eq.(4) to be $\sum_{i=1}^n w_i(a_ib_i)$ with $\bm{w}=[w_1,w_2,\cdots,w_n]^{\rm{T}}$. Thus, we obtain a new objective function as Eq.(5).
\begin{equation}
  \sum_{i=1}^n w_i(a_ib_i) = \bm{u}\bm{X}^{\rm{T}}\mbox{diag}(\bm{w})\bm{Y}\bm{v}
\end{equation}
Furthermore, we also force $\bm{w}$ to be sparse to select a limited number of samples. Finally we propose a $L_0$-regularized SWCCA ($L_0$-SWCCA) model as Eq.(6).
\begin{equation}
  \begin{array}{rl}
    \max_{\bm{u},\bm{v},\bm{w}} & \bm{u}^{\rm{T}}\bm{X}^{\rm{T}}\mbox{diag}(\bm{w})\bm{Y}\bm{v}\\
    \mbox{s.t.} & \|\bm{u}\|_0\leq k_u, \|\bm{v}\|_0\leq k_v, \|\bm{w}\|_0\leq k_w\\
                & \bm{u}^{\rm{T}}\bm{u} = \bm{v}^{\rm{T}}\bm{v} = \bm{w}^{\rm{T}}\bm{w} = 1
  \end{array}
\end{equation}
where $\mbox{diag}(\bm{w})$ is a diagonal matrix and $\mbox{diag}(\bm{w})_{ii} = w_i$. If $\mbox{diag}(\bm{w}) = \frac{1}{\sqrt{n}}\bm{I}$, then $L_0$-SWCCA reduces to $L_0$-SCCA.

\section{Optimization}
In this section, we design an alternating iterative algorithm to solve (6) by using a sparse projection strategy. We start with the sparse projection problem corresponding to the sub-problem of (6) with fixed $\bm{v}$ and $\bm{w}$ as Eq.(7).
\begin{equation}
  \begin{array}{rl}
    \max_{\bm{u}} & \bm{u}^{\rm{T}}\bm{z} \\
    \mbox{s.t.}& \bm{u}^{\rm{T}}\bm{u} =  1, \|\bm{u}\|_0 \leq k
  \end{array}
\end{equation}
For a given column vector $\bm{z} \in \mathbb{R}^{p \times 1}$ and $k\leq p$, we define a sparse project operator $\Pi(\cdot,k)$ as Eq.(8).
\begin{eqnarray}
  [\Pi(\bm{z},k)]_i =
  \begin{cases}
    z_i, &\mbox{if}~i \in \mbox{support}(\bm{z},k)\cr
    0, &\mbox{otherwise}
  \end{cases}
\end{eqnarray}
where $\mbox{support}(\bm{z},k)$ is defined as a set of indexes corresponding to the largest $k$ absolute values of $\bm{z}$. For example, if $\bm{z}=[-5, 3, 5, 2, -1]^{\rm{T}}$, then $\Pi(\bm{z},3) = [-5, 3, 5, 0, 0]^{\rm{T}}$.

{\bf Theorem 1}\quad  The solution of problem (7) is
\begin{equation}
  \bm{u}^* =  \frac{\Pi(\bm{z},k)}{\|\Pi(\bm{z},k)\|_2}
\end{equation}

Note that $\|\cdot\|_2$ denotes the Euclidean norm. We can prove the Theorem~1 by contradiction (we omit the proof here). Based on Theorem~1, we design an alternating iterative approach to solve Eq.(6).

i) Optimizing $\bm{u}$ with fixed $\bm{v}$ and $\bm{w}$.
Fix $\bm{v}$ and $\bm{w}$ in Eq.(6), let $\bm{z}_u =\bm{X}^{\rm{T}}\mbox{diag}(\bm{w})\bm{Y}\bm{v}$, then Eq.(6) reduces to as Eq.(10).
\begin{equation}
  \begin{array}{rl}
    \max_{\bm{u}} & \bm{u}^{\rm{T}}\bm{z}_u \\
    \mbox{s.t.}& \bm{u}^{\rm{T}}\bm{u} =  1, \|\bm{u}\|_0 \leq k_u
  \end{array}
\end{equation}
Based on the Theorem~1, we obtain the update rule of $\bm{u}$ as Eq.(11).
\begin{equation}
   \bm{u} \leftarrow \frac{\Pi(\bm{z}_u,k_u)}{\|\Pi(\bm{z}_u,k_u)\|_2}
\end{equation}

ii) Optimizing $\bm{v}$ with fixed $\bm{u}$ and $\bm{w}$.
Fix $\bm{u}$ and $\bm{w}$ in Eq.(6), let $\bm{z}_v=\bm{Y}^{\rm{T}}\mbox{diag}(\bm{w})\bm{X}\bm{u}$, then Eq.(6) reduces to as Eq.(12).
\begin{equation}
  \begin{array}{rl}
    \max_{\bm{v}} & \bm{v}^{\rm{T}}\bm{z}_v \\
    \mbox{s.t.}& \bm{v}^{\rm{T}}\bm{v} =  1, \|\bm{v}\|_0 \leq k_v
  \end{array}
\end{equation}
Similarly, we obtain the update rule of $\bm{v}$ as Eq.(13).
\begin{equation}
   \bm{v} \leftarrow \frac{\Pi(\bm{z}_v,k_v)}{\|\Pi(\bm{z}_v,k_v)\|_2}
\end{equation}

iii) Optimizing $\bm{w}$ with fixed $\bm{u}$ and $\bm{v}$.
Fix $\bm{u}$ and $\bm{v}$ in Eq.(6), then Eq.(6) reduces to as Eq.(14).
\begin{equation}
  \begin{array}{rl}
    \max_{\bm{w}} & \bm{u}^{\rm{T}}\bm{X}^{\rm{T}}\mbox{diag}(\bm{w})\bm{Y}\bm{v}\\
    \mbox{s.t.}& \bm{w}^{\rm{T}}\bm{w} =  1, \|\bm{w}\|_0 \leq k_u
  \end{array}
\end{equation}
Let $\bm{t}_1 = \bm{Xu}$, $\bm{t}_2 = \bm{Yu}$ and $\bm{z}_w = \bm{t}_1\odot \bm{t}_2$ where `$\odot$' denotes point multiplication which is equivalent to `.*' in Matlab, then we have
$\bm{u}^{\rm{T}}\bm{X}^{\rm{T}}\mbox{diag}(\bm{w})\bm{Y}\bm{v} =\bm{t}_1\mbox{diag}(\bm{w})\bm{t}_2 = (\bm{t}_1\odot\bm{w})^{\rm{T}}\bm{t}_2 = \bm{w}^{\rm{T}}(\bm{t}_1\odot\bm{t}_2) = \bm{w}^{\rm{T}}\bm{z}_w$. Thus, problem (14) reduces to as Eq.(15).
\begin{equation}
  \begin{array}{rl}
    \max_{\bm{w}} & \bm{w}^{\rm{T}}\bm{z}_w\\
    \mbox{s.t.}& \bm{w}^{\rm{T}}\bm{w} =  1, \|\bm{w}\|_0 \leq k_w
  \end{array}
\end{equation}
Similarly, we obtain the update rule of $\bm{w}$ as Eq.(16).
\begin{equation}
   \bm{w} \leftarrow \frac{\Pi(\bm{z}_w,k_w)}{\|\Pi(\bm{z}_w,k_w)\|_2}
\end{equation}
Finally, combining (11), (13) and (16), we propose the following alternating iterative algorithm to solve problem (6) as Algorithm 1.

\begin{figure*}[htbp]
  \centering
  \includegraphics[width=1\linewidth]{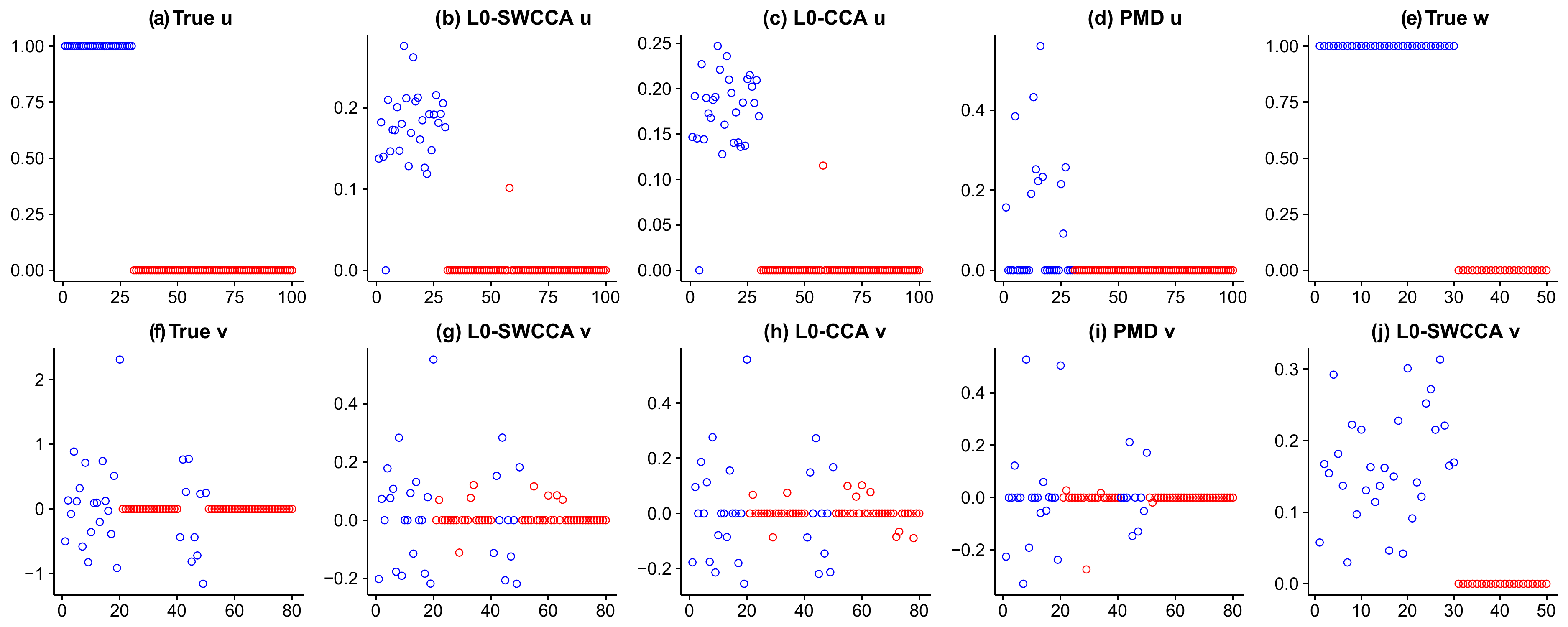}
  \caption{Results of the synthetic data 1. (a), (f) and (e) denote true $\bm{u}$, $\bm{v}$ and $\bm{w}$; (b), (g) and (j) denote estimated $\bm{u}$, $\bm{v}$ and $\bm{w}$ by $L_0$-SWCCA; (c) and (h) denote estimated $\bm{u}$ and $\bm{v}$ by $L_0$-SCCA; (d) and (i) denote stimated $\bm{u}$ and $\bm{v}$ by PMD.}
\end{figure*}

\begin{algorithm}[h]
\caption{\textbf{$\bm{L_0}$-SWCCA}.} \label{alg:Framwork1}
\begin{algorithmic}[1]
\REQUIRE $\bm{X}\in \mathbb{R}^{n\times p}$, $\bm{Y}\in \mathbb{R}^{n\times q}$, $k_u$, $k_v$ and $k_w$.
\ENSURE $\bm{u}$, $\bm{v}$, and $\bm{w}$.
\STATE Initial  $\bm{u}$, $\bm{v}$, and $\bm{w}$
\REPEAT
\STATE Update $\bm{u}$ according to Eq.(11)
\STATE Update $\bm{v}$ according to Eq.(13)
\STATE Update $\bm{w}$ according to Eq.(16)
\UNTIL convergence of $\bm{u}$, $\bm{v}$, and $\bm{w}$.
\end{algorithmic}
\end{algorithm}

{\bf Terminating Condition:} We can set different stop conditions to control the iterations. For example, the update length of $\bm{u}$, $\bm{v}$, and $\bm{w}$ are smaller than a given threshold (\emph{i.e.}, $\|\bm{u}^k - \bm{u}^{k+1}\|_2^2<10^{-6}$, $\|\bm{v}^k - \bm{v}^{k+1}\|_2^2<10^{-6}$ and $\|\bm{w}^k - \bm{w}^{k+1}\|_2^2<10^{-6}$), or the maximum number of iterations is a given number (\emph{e.g.}, 1000), or the change of objective value is less than a give threshold.

{\bf Computation Complexity:} The complexity of matrix multiplication with one $n\times p$ matrix and another one $n \times q$ is $\mathcal{O}(npq)$. To reduce the computational complexity of $\bm{X}^{\rm{T}}\mbox{diag}(\bm{w})\bm{Y}\bm{v}$, we note that $\bm{X}^{\rm{T}}\mbox{diag}(\bm{w})\bm{Y}\bm{v} = \bm{X}^{\rm{T}}(\mbox{diag}(\bm{w})\bm{Y}\bm{v}) = \bm{X}^{\rm{T}}[(\bm{Y}\bm{v})\odot\bm{w}]$. Let $\bm{t}_1 = \bm{Y}\bm{v}$, $\bm{t}_2 = \bm{t}_1\odot\bm{w}$ and $\bm{t}_3 = \bm{X}^{\rm{T}}\bm{t}_2$. Thus, the complexity of $\bm{X}^{\rm{T}}\mbox{diag}(\bm{w})\bm{Y}\bm{v}$ is $\mathcal{O}(nq+n+np)$. Similarly, we can see that the complexity of $\bm{Y}^{\rm{T}}\mbox{diag}(\bm{w})\bm{X}\bm{u}$ is $\mathcal{O}(np+n+nq)$, and the complexity of $(\bm{Xu})\odot(\bm{Yv})$ is $\mathcal{O}(nq+np+n)$.
In Algorithm 1, we need to obtain the largest $k$ absolute values of a given vector $\bm{z}$ of size $p\times 1$ [\emph{i.e.}, $\Pi(\bm{z},k)$]. We adopt a linear time selection algorithm called Quick select (QS) algorithm to compute $\Pi(\bm{z},k)$, which applies a divide and conquer strategy, and the average time complexity of QS algorithm is $\mathcal{O}(p)$. Thus, the entire time complexity of Algorithm 1 is $\mathcal{O}(Tnp+Tnq)$, where $T$ is the number of iterations for convergence. In general, $T$ is a small number.

{\bf Convergence Analysis:} Similar with Theorem 1 in Ref. \cite{sun2015multi}, also see e.g., \cite{bolte2014proximal}, we can prove that Algorithm 1 converges globally to a critical point (we omit the proof here).
\section{Experiments}
\subsection{Synthetic data 1}
Here we generate the first synthetic data matrices $\bm{X}$ and $\bm{Y}$ with $n = 50$, $p = 100$ and  $q = 80$ using the following two steps:

Step 1: Generate two canonical vectors $\bm{u}$, $\bm{v}$ and a weighted vector $\bm{w}$ as Eq.(17).
\begin{equation}
  \begin{array}{rl}
    \bm{u}~~= & [r(1,30),r(0,70)]^{\rm{T}}\\
    \bm{v}~~= & [N(20),r(0,20),N(10),r(0,30)]^{\rm{T}}\\
    \bm{w}~~= & [r(1,30),r(0,20)]^{\rm{T}}\\
  \end{array}
\end{equation}
where $r(a,n)$ denotes a row vector of size $n$, whose elements are equal to $a$, $N(m)$ denotes a row vector of size $m$, whose elements are randomly sampled from a standard normal distribution.

Step 2: Generate two input matrices $\bm{X}$ and $\bm{Y}$ as Eq.(18).
\begin{equation}
  \begin{array}{rl}
    \bm{X}~~= & \bm{w}\bm{u}^{\rm{T}} + \bm{\epsilon}_x \\
    \bm{Y}~~= & \bm{w}\bm{v}^{\rm{T}} + \bm{\epsilon}_y
  \end{array}
\end{equation}
where the elements of $\bm{\epsilon}_x$ and $\bm{\epsilon}_y$ are randomly sampled from a standard normal distribution.

We evaluate the performance of $L_0$-SWCCA with the above synthetic data and compare its performance with the typical sparse CCA, including $L_0$-SCCA \cite{asteris2016simple} and PMD \cite{witten2009penalized} with $L_1$-penalty. For comparison,  we set parameters $k_u = 30$, $k_v = 30$ and $k_w = 30$ for $L_0$-SWCCA; $k_u = 30$, $k_v = 30$ for $L_0$-SCCA; $c_1 = \frac{30}{100}\sqrt{p}$ and $c_2 = \frac{30}{80}\sqrt{q}$ for PMD. Note that $c_1 = c\sqrt{p}$ and $c_2 = c\sqrt{q}$ where $c \in (0, 1)$ for PMD are to approximately control the sparse proportion of the canonical vectors ($\bm{u}$ and $\bm{v}$).

The true and estimated patterns for $\bm{u}$, $\bm{v}$ and $\bm{w}$ in the synthetic data 1 are shown in Fig.2. Compared to PMD, $L_0$-SWCCA and $L_0$-SCCA does fairly well for identifying the local non-zero pattern of the underlying factors (\emph{i.e.}, $\bm{u}$ and $\bm{v}$). However, the two traditional SCCA methods ($L_0$-SCCA and PMD) do not recognize the difference between samples and remove the noisy samples. Interestingly, $L_0$-SWCCA not only discovers the true patterns for $\bm{u}$, $\bm{v}$ (Fig.2(b) and (g)), but also identifies the true non-zero characteristics of samples ($\bm{w}$) (Fig.2(e)). Furthermore, to assess our approach is indeed able to find a greater correlation level between two input matrices, we define the correlation criterion as Eq.(19).
\begin{equation}
\rho = \mbox{cor}((\bm{X}\bm{u})\odot\bm{w}, (\bm{Y}\bm{v})\odot\bm{w})
\end{equation}
where $\mbox{cor}(\cdot)$ is a function to calculate the correlation coefficient of the two vectors. For comparison, we set $\bm{w} = [1, \cdots, 1]^{\rm{T}}$ for $L_0$-SCCA and PMD to compute the correlation criterion.
$L_0$-SWCCA gets the largest $\rho=0.96$ compared to $L_0$-SCCA with $\rho=0.80$ and PMD with $\rho=0.87$ in the above synthetic data. All results show that $L_0$-SWCCA is more effective to capture the latent patterns of canonical vectors than other methods.
\subsection{Synthetic data 2}
Here we apply another way to generate synthetic data matrices $\bm{X}\in \mathbb{R}^{n\times p}$ and $\bm{Y}\in \mathbb{R}^{n\times q}$ with $n = 50$, $p = 100$, and  $q = 80$. The following three steps are used to generate the second synthetic data matrices $\bm{X}$ and $\bm{Y}$:

Step 1: We first generate two zero matrices as Eq.(20).
\begin{equation}
  \begin{array}{rl}
    \bm{X}~~\leftarrow & matrix(0,nrow=n, ncol=p)\\
    \bm{Y}~~\leftarrow & matrix(0,nrow=n, ncol=q)
  \end{array}
\end{equation}

Step 2: We then update two sub-matrices in the $\bm{X}$ and $\bm{Y}$.
\begin{equation}
  \begin{array}{rl}
    \bm{X}[1:30,1:50]~~\leftarrow & 1\\
    \bm{Y}[1:30,1:40]~~\leftarrow & - 1
  \end{array}
\end{equation}

Step 3: We add the Gaussian noise in $\bm{X}$ and $\bm{Y}$.
\begin{equation}
  \begin{array}{rl}
    \bm{X}~~\leftarrow &  \bm{X} +  \bm{\epsilon}_x\\
    \bm{Y}~~\leftarrow &  \bm{Y} +  \bm{\epsilon}_y
  \end{array}
\end{equation}
\begin{figure*}[htbp]
  \centering
  \includegraphics[width=1\linewidth]{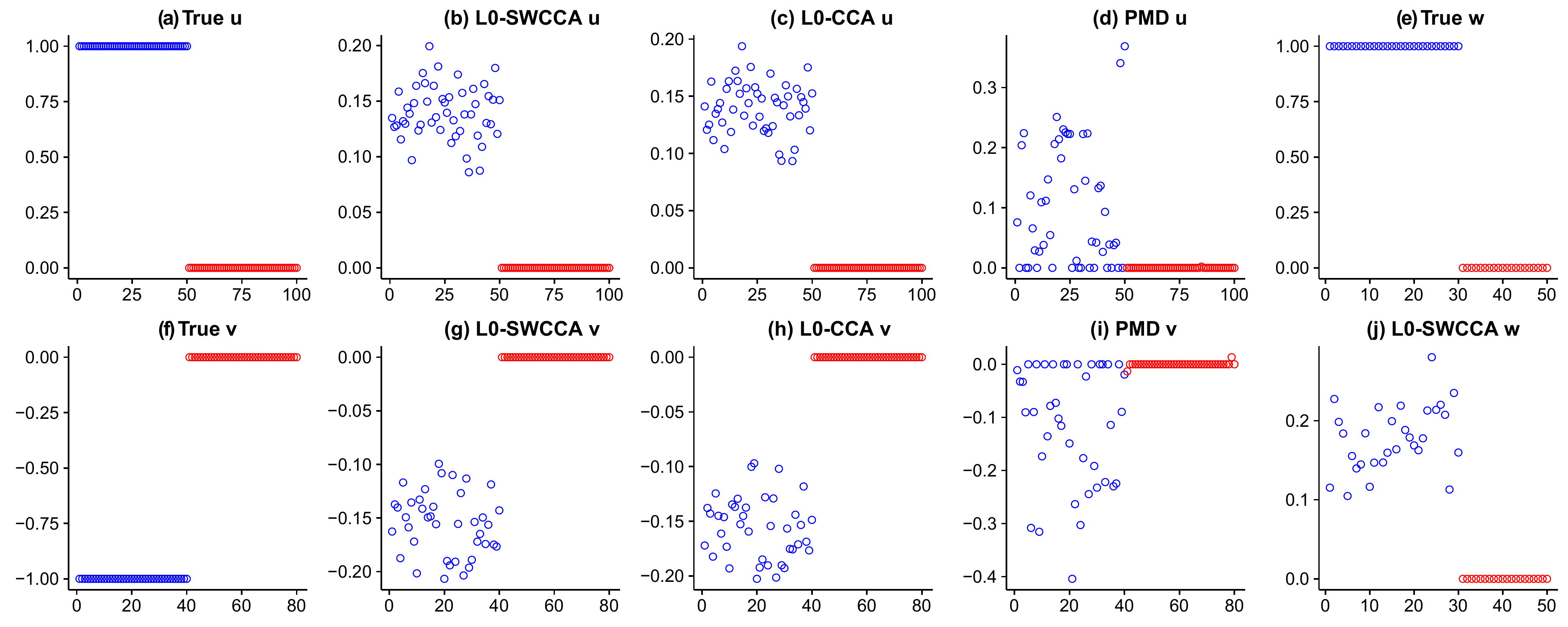}
  \caption{Results of the synthetic data 2. (a), (f) and (e) denote true $\bm{u}$, $\bm{v}$ and $\bm{w}$; (b), (g) and (j) denote estimated $\bm{u}$, $\bm{v}$ and $\bm{w}$ by $L_0$-SWCCA; (c) and (h) denote estimated $\bm{u}$ and $\bm{v}$ by $L_0$-SCCA; (d) and (i) denote estimated $\bm{u}$ and $\bm{v}$ by PMD.}
\end{figure*}

For simplicity and comparison, we can set true $\bm{u} = [r(1,50),~r(0,50)]^{\rm{T}}$, true $\bm{v} = [r(-1,40),~r(0,40)]^{\rm{T}}$ and true $\bm{w} = [r(1,30),r(0,20)]^{\rm{T}}$ to characterize the patterns of $\bm{X}$ and $\bm{Y}$ (Fig.3(a), (f) and (e)). Similarly, we also apply $L_0$-SWCCA, $L_0$-SCCA \cite{asteris2016simple} and PMD \cite{witten2009penalized} to the synthetic data 2. For comparison, we set parameters $k_u = 50$, $k_v = 40$ and $k_w = 30$ for $L_0$-SWCCA; $k_u = 50$, $k_v = 40$ for $L_0$-SCCA; $c_1 = \frac{50}{100}\sqrt{p}$ and $c_2 = \frac{40}{80}\sqrt{q}$ for PMD.

The true and estimated patterns for $\bm{u}$, $\bm{v}$ and $\bm{w}$ are shown in Fig.3. $L_0$-SWCCA and $L_0$-SCCA are superior to PMD about identifying the latent patterns of canonical vectors $\bm{u}$ and $\bm{v}$ (Fig.3(b), (c), (d), (g), (h) and (i)). However $L_0$-SCCA and PMD fail to remove interference samples. Compared to $L_0$-SCCA and PMD, $L_0$-SWCCA can clearly identify the true non-zero characteristics of samples (Fig.3(e)). Similarly, we also compute the correlation criterion based on the formula (19). We find that $L_0$-SWCCA gets the largest correlation $\rho=0.97$ compared to $L_0$-SCCA with $\rho=0.93$ and PMD with $\rho=0.95$. All results show that our method is more effective to capture the latent patterns of canonical vectors than other ones.

\subsection{Breast cancer data}
We first consider a breast cancer dataset \cite{witten2009penalized,chin2006genomic} consisting of gene expression and DNA copy number variation data across 89 cancer samples. Specifically, the gene expression data $\bm{X}$ and the DNA copy number data $\bm{Y}$ are of size $n \times p$ and  $n \times q$ with $n = 89$, $p = 19672$ and $q = 2149$. We apply SWCCA and related ones to this data to identify a gene set whose expression is strongly correlated with copy number changes of some genomic regions.

In PMD \cite{witten2009penalized}, we set $c_1 = c\sqrt{p}$ and $c_2 = c\sqrt{q}$, where $c \in (0, 1)$ is to approximately control the sparse ratio of canonical vectors. We ensure that the canonical vectors ($\bm{u}$ and $\bm{v}$) extracted by the three methods (PMD, $L_0$-SCCA, and $L_0$-SWCCA) have the same sparsity level for comparison. We first apply PMD in the breast cancer data to obtain two sparse canonical vectors $\bm{u}$ and $\bm{v}$ for each given $c \in (0,1)$. Then, we compute the number of nonzero elements in the above extracted $\bm{u}$ and $\bm{v}$, denoted as $N_u$ and $N_v$. Finally, we set $k_u = N_u$, $k_v = N_v$ in $L_0$-SCCA and $L_0$-SWCCA, and set $k_w =  53 \approx 0.6\times89$ in $L_0$-SWCCA to identify the sample loading $\bm{w}$ with sparse ratio $60\%$.

We adopt two criteria: correlation level defined in formula~(19) and objective value defined in Eq.(5) for comparison. Here we consider different $c$ values (\emph{i.e.}, $0.1,0.2,0.3,0.4,0.5,0.6,0.7$) to  control the different sparse ratio of canonical vectors. We find that, compared to PMD and $L_0$-SCCA, $L_0$-SWCCA does obtain higher correlation level and objective value for all cases (Table 1). Since the `breast cancer data' did not collect any clinical information of patients, it is very difficult to study the specific characteristics of these selected samples. To this end, we also apply our method to another biological data with more detailed clinical information.

{\tabcolsep=2.7pt \footnotesize
\begin{center}
\begin{tabular}{|c|c|c|c|c|c|c|c|}
    \multicolumn{8}{c}{\bf Table 1.~Results on  Correlation level (CL) and Objective}\\

    \multicolumn{8}{c}{\bf value (OV) for different c values.}\\
    \hline
    {\bf CL}   & c=0.1 & c=0.2 & c=0.3 & c=0.4 & c=0.5 & c=0.6 & c=0.7\\
	\hline
    PMD & 0.77 & 0.81 & 0.85 & 0.86 & 0.86 & 0.85 & 0.83\\
	\hline
    $L_0$-SCCA& 0.88 & 0.84 & 0.86 & 0.85 & 0.84 & 0.82 & 0.81\\
	\hline
    $L_0$-SWCCA & 0.99 & 0.98 & 1.00 & 1.00 & 0.99 & 0.97 & 0.99\\
	\hline
     \hline
    {\bf OV}  & c=0.1 & c=0.2 & c=0.3 & c=0.4 & c=0.5 & c=0.6 & c=0.7\\
	\hline
    PMD &253  & 768  & 1390 & 2020 & 2589 & 3056  & 3392\\
	\hline
    $L_0$-SCCA   &371  & 1066 & 1824 & 2467 & 3014 & 3360  & 3520\\
	\hline
    $L_0$-SWCCA &1570 & 3046 & 4960 & 6576 & 7860 & 10375 & 10692\\
	\hline
\end{tabular}
\end{center}}
\subsection{TCGA BLCA data}
Recently, it is a hot topic to study microRNA (miRNA) and gene regulatory relationship from matched miRNA and gene expression data \cite{min2015novel,zhang2011novel}. Here, we apply SWCCA onto the bladder urothelial carcinoma (BLCA) miRNA and gene expression data across 405 patients from TCGA (\url{https://cancergenome.nih.gov/}) to identify a subtype-specific miRNA-gene co-correlation module. To remove some noise miRNAs and genes, we first adapt standard deviation method to extract 200 largest variance  miRNAs and 5000 largest variance genes for further analysis. Finally, we obtain a miRNA expression matrix $\bm{X}\in \mathbb{R}^{405\times200}$, which is standardized for each miRNA, and a gene expression matrix $\bm{Y}\in \mathbb{R}^{405\times5000}$, which is standardized for ecah gene. We apply $L_0$-SWCCA onto BLCA data with $k_u = 10$, $k_v = 200$ and $k_w = 203$ to identify a miRNA set with 10 miRNAs and a gene set with 200 genes and a sample set with 203 patients. We also apply PMD with $c_1=(10/200)\sqrt{p}$, $c_2=(200/5000)\sqrt{q}$ and $L_0$-SCCA with $k_u = 10$ and $k_v = 200$ onto BLCA data for comparison.

\begin{figure}[h]
  \centering
  \includegraphics[width=1\linewidth]{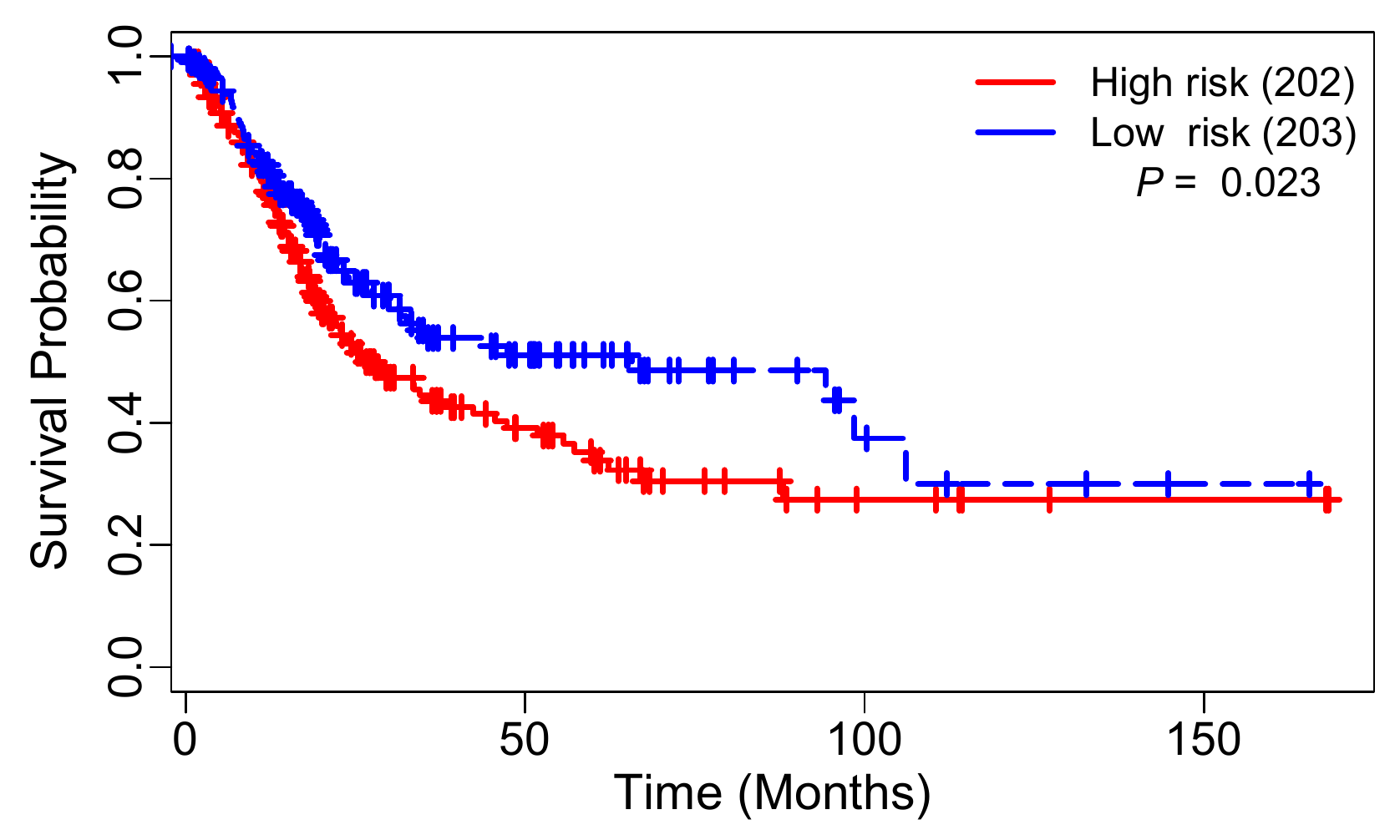}
  \caption{Kaplan-Meier survival analysis between the selected patients and the remaining patents based on the $\bm{w}$ estimated by $L_0$-SWCCA. $P$-value is calculated by log-rank test.}
\end{figure}

Similarly, $L_0$-SWCCA obtains the largest correlation level (CL) and objective value (OV) than others ones [(CL, OV): (0.98, 1210) for $L_0$-SWCCA, (0.84, 346) for PMD, (0.86,469) for $L_0$-SCCA], respectively. More importantly, we also analyze characteristics of these selected patients by $L_0$-SWCCA. We find that it is significantly different with respect to patient survival time between the selected 203 patients and the remaining 202 patients with $p$-value $=0.023$ (Fig.4). These results imply that $L_0$-SWCCA can be used to discover BLCA subtype-specific miNRA-gene co-correlation modules.

Furthermore, we also assess whether these identified genes by $L_0$-SWCCA are biologically relevant with BLCA. DAVID (\url{https://david.ncifcrf.gov/}) is used to perform the Gene Ontology (GO) biological processes (BPs) and KEGG pathways enrichment analysis. Several significantly enriched  GO BPs and pathways relating to BLCA are discovered including \emph{GO:0008544:epidermis development} (B-H adjusted $P$-value $=$ 1.1E-12), \emph{hsa00591:Linoleic acid metabolism} (B-H adjusted $p$-value $=$ 4.8E-3), \emph{hsa00590:Arachidonic acid metabolism} (B-H adjusted $p$-value $=$ 3.6E-3) and \emph{hsa00601:Glycosphingolipid biosynthesis-lacto and neolacto series} (B-H adjusted $p$-value $=$ 2.6E-2). Finally, we also examine whether the identified miRNAs by $L_0$-SWCCA are associated with BLCA. Interestingly, in the identified 10 miRNAs by $L_0$-SWCCA, we find that there are six miRNAs (including \emph{hsa-miR-200a-3p, hsa-miR-200b-5p, hsa-miR-200b-3p, hsa-miR-200a-5p, hsa-miR-200c-3p and hsa-miR-200c-5p}) belonging to miR-200 family. Notably, several studies \cite{wiklund2011coordinated,cheng2016mir} have also reported miR-200 family plays key roles in BLCA. All these results imply that the identified miNRA-gene module by $L_0$-SWCCA may help us to find new therapeutic strategy for BLCA.

\section{Extensions}
\subsection{SWCCA with generalized penalties}
We first consider a general regularized SWCCA framework as Eq.(23).
\begin{equation}
  \begin{array}{rl}
    \max_{\bm{u},\bm{v},\bm{w}} & \bm{u}^{\rm{T}}\bm{X}^{\rm{T}}\mbox{diag}(\bm{w})\bm{Y}\bm{v} \\
                         & - \mathcal{R}_u(\bm{u}) - \mathcal{R}_v(\bm{v})-\mathcal{R}_w(\bm{w})\\
    \mbox{s.t.} & \bm{u}^{\rm{T}}\bm{u} = 1,\bm{v}^{\rm{T}}\bm{v} = 1, \bm{w}^{\rm{T}}\bm{w} = 1
  \end{array}
\end{equation}
where $\mathcal{R}_u(\cdot)$, $\mathcal{R}_v(\cdot)$, $\mathcal{R}_w(\cdot)$ are three regularized functions. For different prior knowledge, we can use different sparsity inducing penalties.

\subsubsection{LASSO regularized SWCCA}
If $\mathcal{R}_u(\bm{u})= \lambda_u\|\bm{u}\|_1$, $\mathcal{R}_v(\bm{v})= \lambda_v\|\bm{v}\|_1$, and $\mathcal{R}_w(\bm{w})= \lambda_w\|\bm{w}\|_1$. We obtain a $L_1$ (Lasso) regularized SWCCA ($L_1$-SWCCA).
Similar to solve Eq.(6), we only need to solve the following problem to solve $L_1$-SWCCA as Eq.(24).
\begin{equation}
  \begin{array}{rl}
    \min_{\bm{u}}&-\bm{u}^{\rm{T}}\bm{z} + \lambda_u\|\bm{u}\|_1\\
    \mbox{s.t.}& \bm{u}^{\rm{T}}\bm{u} =  1
  \end{array}
\end{equation}
where $\bm{z} = \bm{X}^{\rm{T}}\mbox{diag}(\bm{w})\bm{Y}\bm{v}$.
We first replace the constraint $\bm{u}^{\rm{T}}\bm{u} = 1$ with $\bm{u}^{\rm{T}}\bm{u} \leq 1$ and obtain the following the problem as Eq.(25).
\begin{equation}
  \begin{array}{rl}
    \min_{\bm{u}}&-\bm{u}^{\rm{T}}\bm{z} + \lambda_u\|\bm{u}\|_1\\
    \mbox{s.t.}& \bm{u}^{\rm{T}}\bm{u} \leq 1
  \end{array}
\end{equation}
It is easy to see that problem (25) is equivalent to (24). Thus, we can obtain its Lagrangian form as Eq.(26).
\begin{equation}
\mathcal{L}(\bm{u},\lambda_u, \eta_u) = -\bm{u}^{\rm{T}}\bm{z} + \lambda_u\|\bm{u}\|_1 + \eta_u(\bm{u}^{\rm{T}}\bm{u}-1)
\end{equation}
Thus, we can use a coordinate descent method to minimize Eq.(26) and obtain the following update rule of $\bm{u}$ as Eq.(27).
\begin{equation}
\bm{u} = \frac{\mathcal{S}_{\lambda_u}(\bm{z})}{\|\mathcal{S}_{\lambda_u}(\bm{z})\|_2}
\end{equation}
where $\mathcal{S}_{\lambda_u}(\cdot)$ is a soft thresholding operator and $\mathcal{S}_{\lambda_u}(z_i)=\mbox{sign}(|z_i|-\lambda_u)_+$. Based on the above, an alternating iterative strategy can be used to solve $L_1$-SWCCA.
\subsubsection{Group LASSO regularized SWCCA}
If $\mathcal{R}_u(\bm{u})= \lambda_u \sum_l\|\bm{u}^{(l)}\|_2$, $\mathcal{R}_v(\bm{v})= \lambda_v \sum_l\|\bm{v}^{(l)}\|_2$ and $\mathcal{R}_w(\bm{w})=  \lambda_w \sum_l\|\bm{w}^{(l)}\|_2$. Problem (24) reduces to $L_{2,1}$-regularized SWCCA ($L_{2,1}$-SWCCA). Similarly, we should solve the following projection problem as Eq.(28).
\begin{equation}
  \begin{array}{rl}
    \min_{\bm{u}}&-\bm{u}^{\rm{T}}\bm{z} + \lambda_u\sum\limits_l\|\bm{u}^{(l)}\|_2\\
    \mbox{s.t.}& \bm{u}^{\rm{T}}\bm{u} \leq 1
  \end{array}
\end{equation}
Thus, we obtain its Lagrangian form as Eq.(29).
\begin{equation}
\mathcal{L}(\bm{u},\lambda_u, \eta_u) = -\bm{u}^{\rm{T}}\bm{z} + \lambda_u\sum_l\|\bm{u}^{(l)}\|_2 + \eta_u(\bm{u}^{\rm{T}}\bm{u}-1)
\end{equation}
where $\bm{u}^{(l)}$ is the $l$th group of $\bm{u}$. We adopt a block-coordinate descent method \cite{tseng2001convergence} to solve it and obtain the learning rule of $\bm{u}^{(l)}$ ($l = 1,\cdots, L$) as Eq.(30).
\begin{eqnarray}
  \bm{u}^{(l)}=
  \begin{cases}
    \frac{1}{2\eta_u }\bm{z}^{(l)} (1- \frac{\lambda_u}{\|\bm{z}^{(l)}\|_2} ), &\mbox{if}~\|\bm{z}^{(l)}\|_2 > \lambda_u,\cr
    \bm{0}, &\mbox{otherwise}.
  \end{cases}
\end{eqnarray}
By cyclically applying the above updates, we can minimize Eq.(29). Thus, an alternating iterative strategy can be used to solve $L_{2,1}$-SWCCA.

\subsection{Multi-view sparse weighted CCA}
In various scientific fields, multiple view data (more than two views) can be available from multiple sources or diverse feature subsets. For example, multiple high-throughput molecular profiling data by omics technologies can be produced for the same individuals in bioinformatics \cite{li2012identifying,min2015novel,sun2015multi}. Integrating these data together can significantly increase the power of pattern discovery and individual classification.

Here we extend SWCCA to Multi-view SWCCA (MSWCCA) model for multi-view data analysis (Fig.5) as follows:
\begin{eqnarray*}
  \begin{array}{rl}
    \max_{\bm{u}_i, \bm{w}} & \bm{w}^{\rm{T}}\big[\bigodot\limits_{i=1}^M(\bm{X}_i\bm{u}_i)\big] - \sum\limits_{i=1}^M\mathcal{R}_{\bm{u}_i}(\bm{u}_i)-\mathcal{R}_w(\bm{w})\\
    \mbox{s.t.} & \bm{w}^{\rm{T}}\bm{w} = 1, \bm{u}_i^{\rm{T}}\bm{u}_i = 1~\mbox{for}~i = 1,\cdots, M
  \end{array}
\end{eqnarray*}
where $\bigodot_{i=1}^M(\bm{X}_i\bm{u}_i) = (\bm{X}_1\bm{u}_1)\odot(\bm{X}_2\bm{u}_2)\cdots\odot(\bm{X}_M\bm{u}_M)$. When $M=2$, we can see that $\bm{w}^{\rm{T}}[(\bm{X}_1\bm{u}_1)\odot(\bm{X}_2\bm{u}_2)] = \bm{u}_1^{\rm{T}}\bm{X}_1^{\rm{T}}\mbox{diag}(\bm{w})\bm{X}_2\bm{u}_1$ and it reduces to SWCCA. So we can solve MSWCCA in a similar manner with SWCCA.

\begin{figure}[h]
  \centering
  \includegraphics[width=1\linewidth]{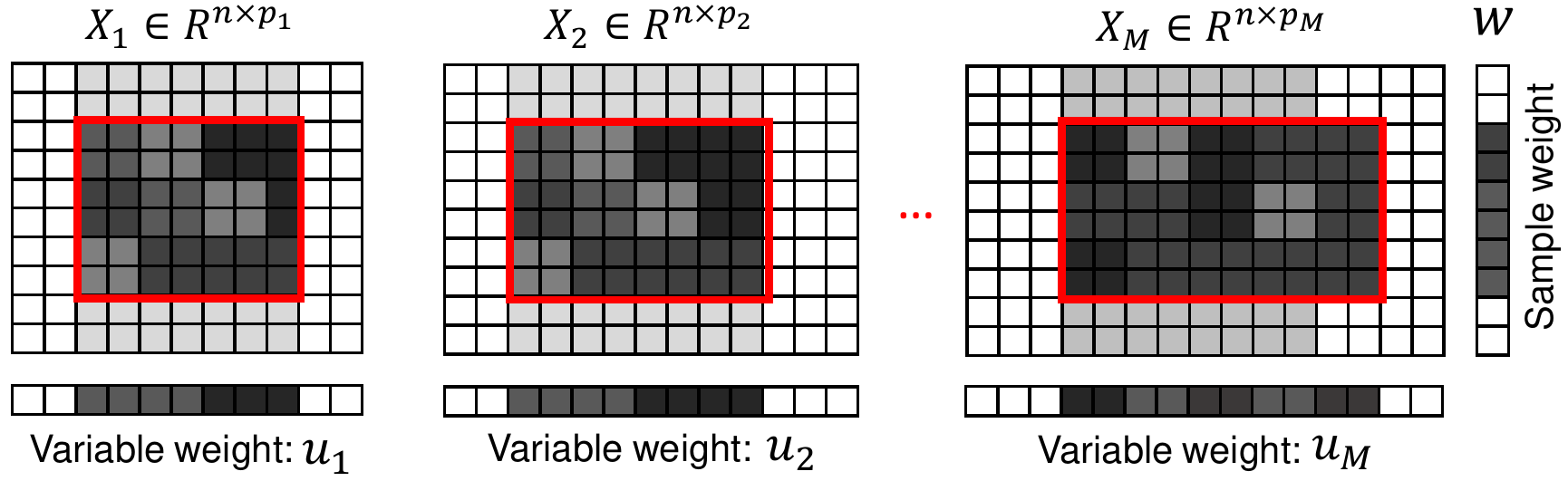}
  \caption{Illustration of the multi-view sparse weighted CCA designed for integrating multiple data matrices.}
\end{figure}
\section{Conclusion}
In this paper, we propose a sparse weighted CCA framework. Compared to SCCA, SWCCA can reveal that the selected variables are only strongly related to a subset of samples. We develop an efficient alternating iterative algorithm to solve the $L_0$-regularized SWCCA. Our tests using both simulation and biological data show that SWCCA can obtain more reasonable patterns compared to the typical SCCA. Moreover, the key idea of SWCCA is easy to be adapted by other penalties like LASSO and Group LASSO. Lastly, we extend SWCCA to MSWCCA for multi-view situation with multiple data matrices.

\section*{Acknowledgment}
Shihua Zhang and Juan Liu are the corresponding authors of this paper. Wenwen Min would like to thank the support of National Center for Mathematics and Interdisciplinary Sciences, Academy of Mathematics and Systems Science, CAS during his visit.

\small{
\bibliographystyle{named}
\balance
\bibliography{SWCCARef}
}
\end{document}